# Application of k-Means Clustering algorithm for prediction of Students' Academic Performance

Oyelade, O. J
Department of Computer and Information Sciences, College of Science and Technology, Covenant University, Ota, Nigeria.
Ola2000faith@yahoo.co.uk.

Oladipupo, O. O
Department of Computer and Information Sciences, College of Science and Technology, Covenant University, Ota, Nigeria.
frajooye@yahoo.com.

Obagbuwa, I. C
Department of Computer Science Lagos State University, Lagos, Nigeria.
ibidunobagbuwa@yahoo.com

*Abstract*— The ability to monitor the progress of students' academic performance is a critical issue to the academic community of higher learning. A system for analyzing students' results based on cluster analysis and uses standard statistical algorithms to arrange their scores data according to the level of their performance is described. In this paper, we also implemented k-mean clustering algorithm for analyzing students' result data. The model was combined with the deterministic model to analyze the students' results of a private Institution in Nigeria which is a good benchmark to monitor the progression of academic performance of students in higher Institution for the purpose of making an effective decision by the academic planners.

## I. INTRODUCTION

Graded Point Average (GPA) is a commonly used indicator of academic performance. Many Universities set a minimum GPA that should be maintained in order to continue in the degree program. In some University, the minimum GPA requirement set for the students is 1.5. Nonetheless, for any graduate program, a GPA of 3.0 and above is considered an indicator of good academic performance. Therefore, GPA still remains the most common factor used by the academic planners to evaluate progression in an academic environment [1]. Many factors could act as barriers to students attaining and maintaining a high GPA that reflects their overall academic performance during their tenure in University. These factors could be targeted by the faculty members in developing strategies to improve student learning and improve their academic performance by way of monitoring the progression of their performance.

Therefore, performance evaluation is one of the bases to monitor the progression of student performance in higher Institution of learning. Base on this critical issue, grouping of students into different categories according to their performance has become a complicated task. With traditional grouping of students based on their average scores, it is difficult to obtain a comprehensive view of the state of the students' performance and simultaneously discover important details from their time to time performance.

With the help of data mining methods, such as clustering algorithm, it is possible to discover the key characteristics from the students' performance and possibly use those characteristics for future prediction. There have been some promising results from applying k-means clustering algorithm with the Euclidean distance measure, where the distance is computed by finding the square of the distance between each scores, summing the squares and finding the square root of the sum [6].

This paper presents k-means clustering algorithm as a simple and efficient tool to monitor the progression of students' performance in higher institution.

Cluster analysis could be divided into hierarchical clustering and non-hierarchical clustering techniques. Examples of hierarchical techniques are single linkage, complete linkage, average linkage, median, and Ward. Non-hierarchical techniques include k-means, adaptive k-means, k-medoids, and fuzzy clustering. To determine which algorithm is good is a function of the type of data available and the particular purpose of analysis. In more objective way, the stability of clusters can be investigated in simulation studies [4]. The problem of selecting the "best" algorithm/parameter setting is a difficult one. A good clustering algorithm ideally should produce groups with distinct non-overlapping boundaries, although a perfect separation can not typically be achieved in practice. Figure of merit measures (indices) such as the silhouette width [4] or the homogeneity index [5] can be used to evaluate the quality of separation obtained using a clustering algorithm. The concept of stability of a clustering algorithm was considered in [3]. The idea behind this validation approach is that an algorithm should be rewarded for consistency. In this paper, we implemented traditional k-means clustering algorithm [6] and Euclidean distance measure of similarity was chosen to be used in the analysis of the students' scores.

## II. METHODOLOGY

### A. Development of k-mean clustering algorithm

Given a dataset of *n* data points $x_1, x_2, ..., x_n$ such that each data point is in $\mathbf{R}^d$, the problem of finding the minimum variance clustering of the dataset into $k$ clusters is that of finding $k$ points $\{m_j\}$ ($j=1, 2, …, k$) in $\mathbf{R}^d$ such that

$$\frac{1}{n}\sum_{i=1}^{n}\left[\min_j d^2(x_i, m_j)\right] \quad (1)$$

is minimized, where $d(x_i, m_j)$ denotes the Euclidean distance between $x_i$ and $m_j$. The points $\{m_j\}$ ($j=1, 2, …,k$) are known as cluster centroids. The problem in Eq.(1) is to find $k$ cluster centroids, such that the average squared Euclidean distance (mean squared error, MSE) between a data point and its nearest cluster centroid is minimized.

The $k$-means algorithm provides an easy method to implement approximate solution to Eq.(1). The reasons for the popularity of $k$-means are ease and simplicity of implementation, scalability, speed of convergence and adaptability to sparse data.

The $k$-means algorithm can be thought of as a gradient descent procedure, which begins at starting cluster centroids, and iteratively updates these centroids to decrease the objective function in Eq.(1). The $k$-means always converge to a local minimum. The particular local minimum found depends on the starting cluster centroids. The problem of finding the global minimum is NP-complete. The $k$-means algorithm updates cluster centroids till local minimum is found. Fig.1 shows the generalized pseudocodes of $k$-means algorithm; and traditional k-means algorithm is presented in fig. 2 respectively.

Before the $k$-means algorithm converges, distance and centroid calculations are done while loops are executed a number of times, say $l$, where the positive integer $l$ is known as the number of $k$-means iterations. The precise value of $l$ varies depending on the initial starting cluster centroids even on the same dataset. So the computational time complexity of the algorithm is $O(nkl)$, where $n$ is the total number of objects in the dataset, $k$ is the required number of clusters we identified and $l$ is the number of iterations, $k \leq n$, $l \leq n$ [6].

```
Step 1:   Accept the number of clusters to group data into and the
          dataset to cluster as input values

Step 2:   Initialize the first K clusters
          -  Take first k instances or
          -  Take Random sampling of k elements

Step 3:   Calculate the arithmetic means of each cluster formed in
          the dataset.

Step 4:   K-means assigns each record in the dataset to only one of
          the initial clusters
          -  Each record is assigned to the nearest cluster using a
             measure of distance (e.g Euclidean distance).

Step 5:   K-means re-assigns each record in the dataset to the most
          similar cluster and re-calculates the arithmetic mean of all
          the clusters in the dataset.
```

Fig 1:    Generalised Pseudocode of Traditional k-means

```
1    MSE = largenumber;
2    Select initial cluster centroids {mj}j
     K = 1;
3    Do
4        OldMSE = MSE;
5        MSE1 = 0;
6        For j = 1 to k
7            mj = 0; nj = 0;
8        endfor
9        For i = 1 to n
10           For j = 1 to k
11               Compute squared Euclidean
                 distance d²(xi, mj);
12           endfor
13           Find the closest centroid mj to xi;
14           mj = mj + xi; nj = nj+1;
15           MSE1 = MSE1 + d²(xi, mj);
16       endfor
17       For j = 1 to k
18           nj = max(nj, 1); mj = mj/nj;
19       endfor
20       MSE = MSE1;
     while (MSE<OldMSE)
```

**Fig.2: Traditional $k$-means algorithm [6]**

### III. RESULTS

We applied the model on the data set (academic result of one semester) of a university in Nigeria. The result generated is shown in tables 2, 3, and 4, respectively. In table 2, for k = 3; in cluster 1, the cluster size is 25 and the overall performance is 62.22. Also, the cluster sizes and the overall performances for cluster numbers 2 and 3 are 15, 29 and 45.73, and 53.03, respectfully. Similar analyses also hold for tables 3 and 4. The graphs are generated in figures 3, 4 and 5, respectively, where the overall performance is plotted against the cluster size.

Table 5 shows the dimension of the data set (Student's scores) in the form N by M matrices, where N is the rows (# of students) and M is the column (# of courses) offered by each student.

The overall performance is evaluated by applying deterministic model in Eq. 2 [7] where the group assessment in each of the cluster size is evaluated by summing the average of the individual scores in each cluster.

$$\frac{1}{N}\left(\sum_{j=1}^{N}\left(\frac{1}{n}\sum_{i=1}^{n} x_i\right)\right) \quad 2$$

Where

N = the total number of students in a cluster and
n = the dimension of the data

**Table 1: Performance index**

| | |
|---|---|
| 70 and above | Excellent |
| 60-69 | Very Good |
| 50-59 | Good |
| 45-49 | Very Fair |
| 40-45 | Fair |
| Below 45 | Poor |

In Figure 3, the overall performance for cluster size 25 is 62.22% while the overall performance for cluster size 15 is 45.73% and cluster size 29 has the overall performance of 53.03%. This analysis showed that, 25 out of 79 students had a "Very Good" performance (62.22%), while 15 out of 79 students had performance in the region of very "Fair" performance (45.73%) and the remaining 29 students had a "Good" performance (53.03%) as depicted in the performance index in table 1.

Figure 4 shows the trends in performance analysis as follows; overall performance for cluster size 24 is 50.08% while the overall performance for cluster size 16 is 65.00%. Cluster size 30 has the overall performance of 58.89%, while cluster size 09 is 43.65%. The trends in this analysis indicated that, 24 students fall in the region of "Good" performance index in table 1 above (50.08%), while 16 students has performance in the region of "Very Good" performance (65.00%). 30 students has a "Good" performance (58.89%) and 9 students had performance of "Fair" result (43.65%).

In figure 5, the overall performance for cluster size 19 is 49.85%, while the overall performance for cluster size 17 is 60.97%. Cluster size 9 has the overall performance of 43.65%, while the cluster size 14 has overall performance of 64.93% and cluster size 20 has overall performance of 55.79%. This performance analysis indicated that, 19 students crossed over to "Good" performance region (49.85%), while 17 students had "Very Good" performance results (60.97%). 9 students fall in the region of "Fair" performance index (43.65%), 14 students were in the region of "Very Good" performance (64.93%) and the remaining 20 students had "Good" performance (55.79%).

**Table 2: K = 3**

| Cluster # | Cluster size | Overall Performance |
|---|---|---|
| 1 | 25 | 62.22 |
| 2 | 15 | 45.73 |
| 3 | 29 | 53.03 |

**Table 3: K = 4**

| Cluster # | Cluster size | Overall Performance |
|---|---|---|
| 1 | 24 | 50.08 |
| 2 | 16 | 65.00 |
| 3 | 30 | 58.89 |
| 4 | 9 | 43.65 |

**Table 4: K = 5**

| Cluster # | Cluster size | Overall Performance |
|---|---|---|
| 1 | 19 | 49.85 |
| 2 | 17 | 60.97 |
| 3 | 9 | 43.65 |
| 4 | 14 | 64.93 |
| 5 | 20 | 55.79 |

**Table 5: Statistics of the Data used**

| Student's Scores | Number of Students | Dimension(Total number of courses) |
|---|---|---|
| Data | 79 | 9 |

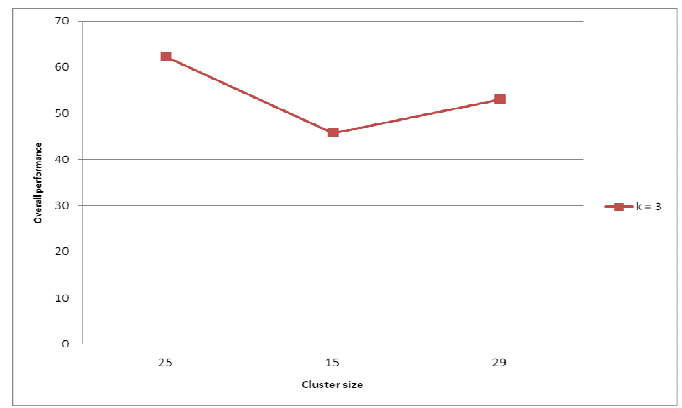

Fig. 3: Overall Performance versus cluster size (# of students) k = 3

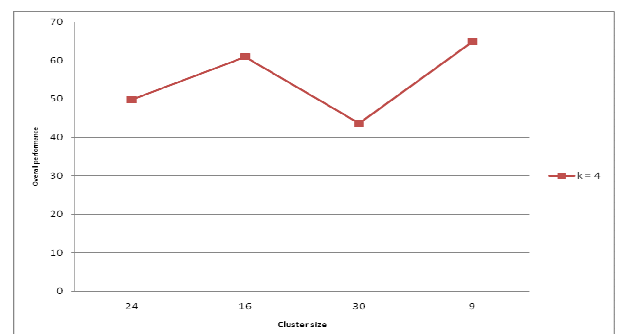

Fig. 4: Overall Performance versus cluster size (# of students) k = 4

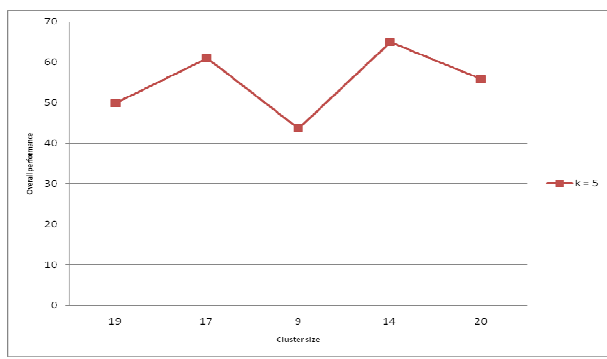

Fig. 5: Overall Performance versus cluster size (# of students) k = 5

## IV. DISCUSSION AND CONCLUSION

In this paper, we provided a simple and qualitative methodology to compare the predictive power of clustering algorithm and the Euclidean distance as a measure of similarity distance. We demonstrated our technique using k-means clustering algorithm [6] and combined with the deterministic model in [7] on a data set of private school results with nine courses offered for that semester for each student for total number of 79 students, and produces the numerical interpretation of the results for the performance evaluation. This model improved on some of the limitations of the existing methods, such as model developed by [7] and [8]. These models applied fuzzy model to predict students' academic performance on two dataset only (English Language and Mathematics) of Secondary Schools results. Also the research work by [9] only provides Data Mining framework for Students' academic performance. The research by [10] used rough Set theory as a classification approach to analyze student data where the Rosetta toolkit was used to evaluate the student data to describe different dependencies between the attributes and the student status where the discovered patterns are explained in plain English.

Therefore, this clustering algorithm serves as a good benchmark to monitor the progression of students' performance in higher institution. It also enhances the decision making by academic planners to monitor the candidates' performance semester by semester by improving on the future academic results in the subsequence academic session.

## ACKNOWLEDGMENT

This work was funded by Covenant University Center for Research and Development. We are grateful to Fahim A. M. and Salem A. M. for their useful materials. We also thank Dr. Obembe for his useful assessment which has improved on the quality of this work.

AUTHORS PROFILE

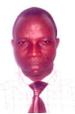

**Oyelade, O. J.:** Received his Bachelor degree in Computer Science with Mathematics(Combined Honour) and M.Sc. in Computer Science from Obafemi Awolowo University, Ile-Ife, Nigeria. He is a Ph.D. Candidate, and a Faculty member in the Department of Computer and Information Sciences, Covenant University, Nigeria. His research interests are in Bioinformatics, Clustering, Fuzzy logic and Algorithms.

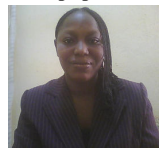

**Oladipupo, O.O.:** Received her Bachelor degree in Computer Science from University of Ilorin, and M.Sc. in Computer Science from Obafemi Awolowo University, Ile-Ife, Nigeria. She is a Ph.D. Candidate, and a Faculty member in the Department of Computer and Information Sciences, Covenant University, Nigeria. Her research interests are in Artificial Intelligent, Data mining and Soft Computing Techniques.

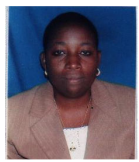

**Obagbuwa, I. C.:** Received her Bachelor degree in Computer Science from University of Ilorin, Ilorin, Nigeria and Master degree (M.Sc.) in Computer Science from University of Port Harcourt, Port Harcourt, Nigeria. She is a Ph.D Candidate in University of Port Harcourt, Port Harcourt, Nigeria in Computer Science. She is a Faculty member in the Department of Computer Sciences, Lagos State University, Ojo – Lagos, Nigeria. Her research interests are in Text segmentation/Automatic Information Extraction, Databases, Document management, Telecommunication and Networking.